\documentclass[]{oppo_arxiv}
\usepackage{threeparttable}
\usepackage{siunitx}

\usepackage[toc,page,header]{appendix}
\usepackage{fancyvrb}
\usepackage{wrapfig}
\usepackage{natbib} 
\setcitestyle{square,numbers,sort&compress}
\usepackage[percent]{overpic}
\usepackage{multibib}
\newcites{case}{Case Study References}
\usepackage{color}
\usepackage{xcolor}
\definecolor{brightube}{rgb}{0.82, 0.62, 0.91}
\definecolor{gentlegreen}{RGB}{34, 139, 34}
\definecolor{softred}{rgb}{0.8, 0.0, 0.0}

\usepackage{times}
\usepackage{latexsym}
\usepackage[T1]{fontenc}
\usepackage[utf8]{inputenc}
\usepackage{microtype}
\usepackage{inconsolata}
\usepackage{graphicx}
\usepackage{hyperref}       
\usepackage{url}            
\usepackage{booktabs}       
\usepackage{amsfonts}       
\usepackage{nicefrac}       
\usepackage{stackengine}
\usepackage{microtype}      
\usepackage{colortbl}
\usepackage{xcolor}
\usepackage{amsmath}
\usepackage{amssymb}
\usepackage{amsthm}
\usepackage{mathrsfs}
\usepackage{pifont}
\usepackage{MnSymbol}
\usepackage{balance}
\usepackage{enumitem}
\usepackage{xcolor}
\usepackage{natbib}
\usepackage{multicol}
\usepackage{xspace}
\usepackage{fancyvrb}
\usepackage{geometry}
\usepackage{tocloft}
\usepackage{tasks}
\usepackage{hyperref}
\usepackage{fontawesome5}
\usepackage{makecell}
\AtBeginDocument{%
  \providecommand\BibTeX{{%
    \normalfont B\kern-0.5em{\scshape i\kern-0.25em b}\kern-0.8em\TeX}}}

\makeatletter
\DeclareRobustCommand\onedot{\futurelet\@let@token\@onedot}
\def\@onedot{\ifx\@let@token.\else.\null\fi}

\usepackage{setspace}
\usepackage{mathtools}

\usepackage{multirow,booktabs}
\usepackage{subcaption}

\title{O-Researcher: An Open Ended Deep Research Model via Multi-Agent Distillation and Agentic RL}

\affiliation{OPPO AI Agent Team}

\abstract{The performance gap between closed-source and open-source large language models (LLMs) is largely attributed to disparities in access to high-quality training data. To bridge this gap, we introduce a novel framework for the automated synthesis of sophisticated, research-grade instructional data. Our approach centers on a multi-agent workflow where collaborative AI agents simulate complex tool-integrated reasoning to generate diverse and high-fidelity data end-to-end. Leveraging this synthesized data, we develop a two-stage training strategy that integrates supervised fine-tuning with a novel reinforcement learning method, designed to maximize model alignment and capability. Extensive experiments demonstrate that our framework empowers open-source models across multiple scales, enabling them to achieve new state-of-the-art performance on the major deep research benchmark. This work provides a scalable and effective pathway for advancing open-source LLMs without relying on proprietary data or models.}

\date{\today}
\checkdata[Open Source]{%
  \href{https://github.com/OPPO-PersonalAI/O-Researcher}{%
    \faGithub \ Code%
  }
  \
  \href{https://huggingface.co/PersonalAILab/O-Researcher-72B-sft}{%
    \raisebox{-.15em}{\includegraphics[height=1em]{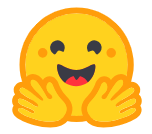}} O-Researcher-SFT%
  }
  \
  \href{https://huggingface.co/PersonalAILab/O-Researcher-72B-rl}{%
    \raisebox{-.15em}{\includegraphics[height=1em]{figures/huggingface.png}} O-Researcher-RL%
  }
  \
}
\correspondence{Wangchunshu Zhou at \email{zhouwangchunshu@oppo.com}}

\begin{document}
\maketitle

\begin{figure}[!h]
    \centering
    \begin{subfigure}[t]{0.55\linewidth}
        \vspace{0pt}
        \centering
        \includegraphics[width=\linewidth]{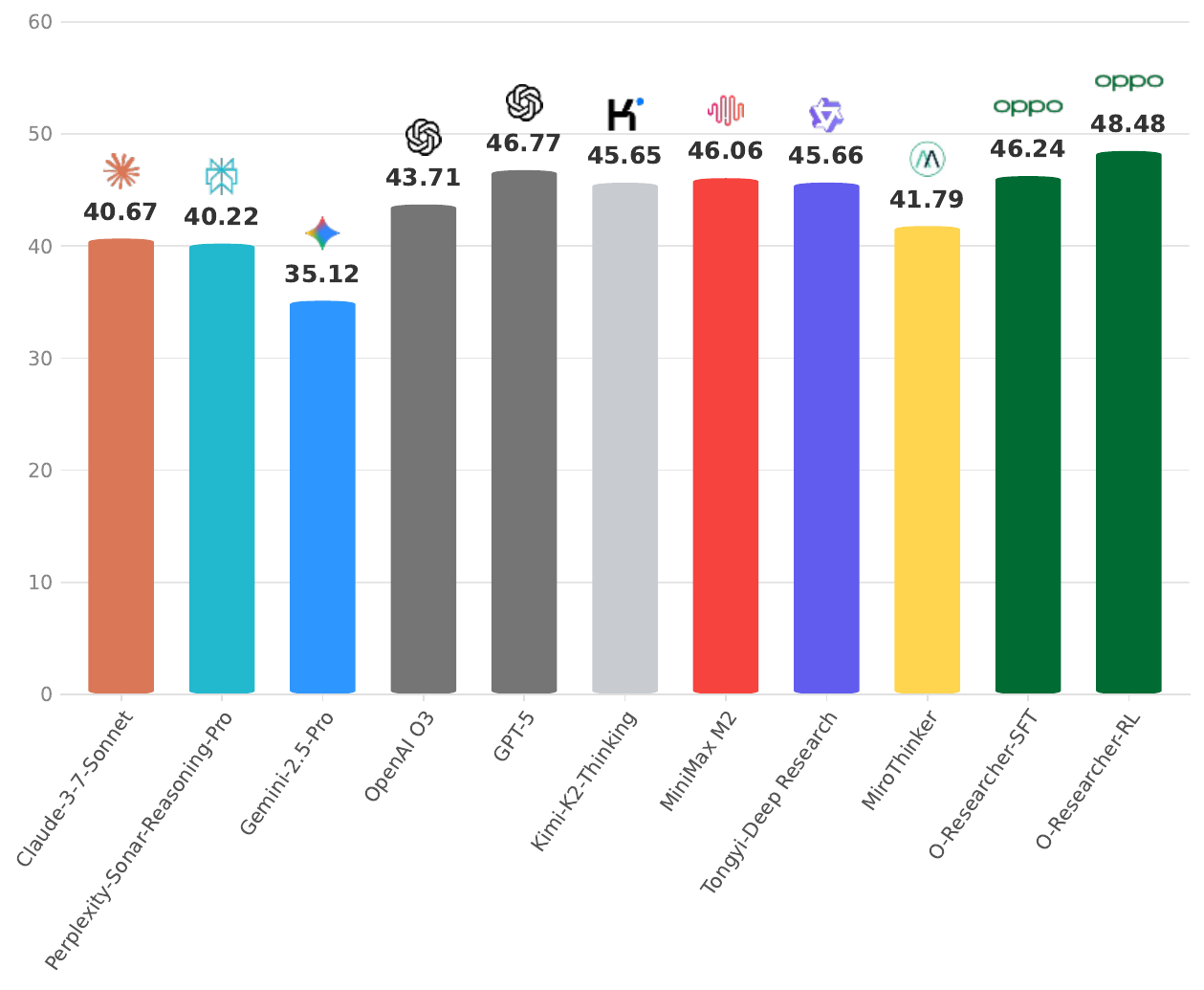}
        \caption{Deep Research Models}
        \label{fig:deep_research_modles}
    \end{subfigure}
    \begin{subfigure}[t]{0.35\linewidth}
        \vspace{0pt}
        \centering
        \includegraphics[width=\linewidth]{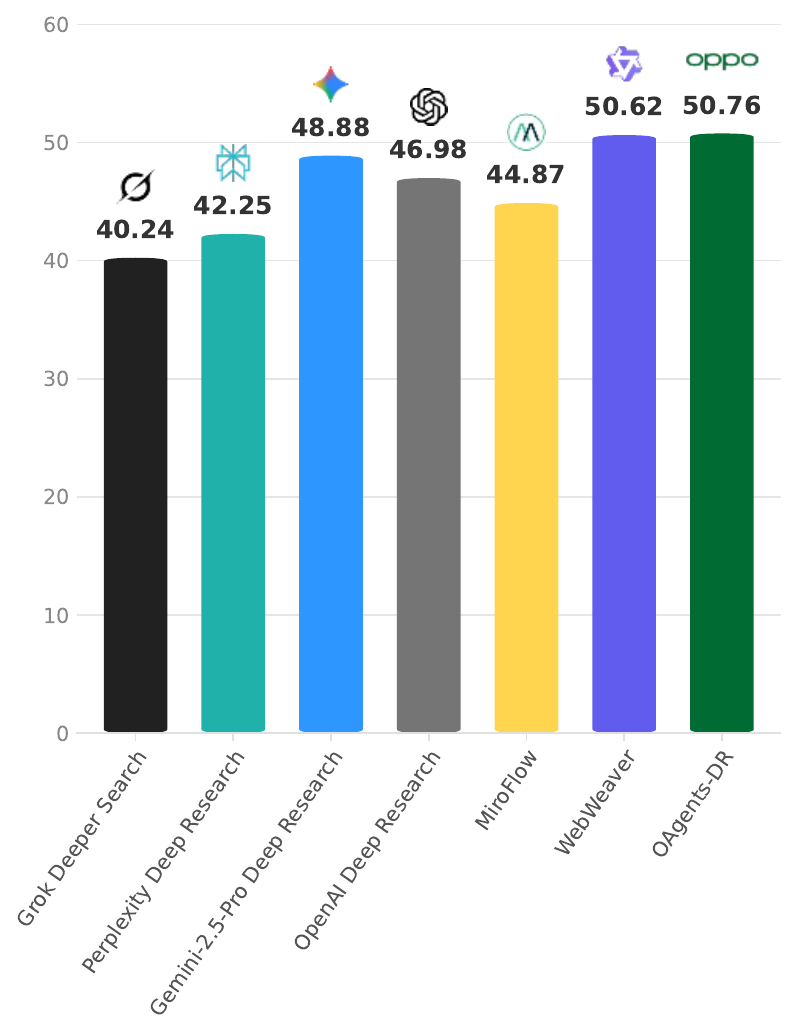}
        \caption{Deep Research Agents}
        \label{fig:deep_research_agents}
    \end{subfigure}
\end{figure}

\section{Introduction}

The rapid evolution of Large Language Models (LLMs) has become a cornerstone of modern artificial intelligence, driving breakthroughs in tasks ranging from natural language understanding to complex reasoning~\cite{dubey2024llama3, openai2024o1}. However, despite these advancements, a persistent challenge within this field is the performance disparity between powerful, closed-source models (e.g., GPT-4o, OpenAI o1~\cite{openai2024o1}) and their open-source counterparts. This gap is often attributed to the vast amounts of proprietary, high-quality training data and immense computational resources available to their developers~\cite{snell2024scaling}. This disparity becomes most apparent in reasoning-intensive tasks that demand extended chains of thought and rigorous problem-solving capabilities~\cite{deepseekai2025deepseekr1}. Consequently, the research community faces a significant bottleneck: how to effectively empower open-source models to achieve state-of-the-art (SOTA) performance without relying on such exclusive advantages.

A critical pathway to overcoming this bottleneck is the generation of high-quality, diverse, and scalable instructional data for supervised fine-tuning (SFT) and reinforcement learning (RL)~\cite{orcutt2024agentinstruct, xu2024magpie}. While existing methods often rely on human annotation or distillation from larger teacher models, these approaches are either prohibitively expensive, limited in scale, or risk inheriting the limitations of the teacher~\cite{gudibande2024falsepromise}. Crucially, standard distillation often captures only the final answer, failing to transfer the intricate ``thought process'' required for complex problem-solving~\cite{deepseekai2025deepseekr1}. Thus, there is a pressing need for an automated, end-to-end framework that can synthesize vast and sophisticated research-grade data to fuel the next generation of open-source LLMs.

In this paper, we propose a novel framework that addresses this fundamental challenge through a multi-agent driven synthetic data generation workflow. Our core insight is that a collaborative ecosystem of specialized AI agents can autonomously simulate complex human reasoning processes—akin to a rigorous peer-review mechanism—to create high-fidelity, multi-turn instruction-response pairs~\cite{shao2024assisting, li2024moreagents}. This synthetic data serves as the foundation for a robust two-stage training strategy, which integrates a novel reinforcement learning method designed to align model outputs with precise quality and correctness metrics~\cite{deepseekai2025deepseekr1, yuan2024selfrewarding}.

Our contributions are summarized as follows:
\begin{itemize}
    \item We raise a novel multi-agent driven workflow to automatically synthesize end-to-end deep research data generation. This system leverages a structured collaboration between multiple LLM agents to decompose, debate, and verify complex tasks, resulting in a scalable pipeline for producing premium training corpora that surpasses the quality of standard model-generated text~\cite{orcutt2024agentinstruct, shao2024assisting}.
    \item Based on the synthesized data from our proposed workflow, we design a two-stage training strategy that incorporates a novel reinforcement learning method. This strategy first employs supervised fine-tuning on the synthetic data to establish a strong knowledge base, followed by a reinforcement learning phase that further refines the model's performance and self-correction capabilities~\cite{deepseekai2025deepseekr1}.
    \item We demonstrate that our method empowers open-source models in multiple sizes to achieve new SOTA performance. Through extensive experiments on major deep research benchmarks, we show that models trained with our framework not only significantly close the gap to closed-source models but also establish new state-of-the-art results for models of comparable size in the open-source domain.
\end{itemize}

\section{Related Work}
Large Reasoning Models (LRMs)~\cite{gpt4_achiam2023gpt, deepseek_guo2025deepseek, qwen3_yang2025qwen3} have achieved impressive results even in challenging domains like math and code. Web agents have also become a significant research direction, with rapid development driven by a series of systems~\cite{li2025websailor, li2025websailorv2, li2025webweaver, wu2025webdancer, li2025webthinker, tao2025webleaper} aiming for web-scale reasoning and information-seeking. Nevertheless, their capabilities remain fundamentally constrained by their internal closed-world knowledge boundaries~\citep{survey2_zhang2025deep}. To mitigate these limitations, agentic systems~\cite{autogpt_yang2023auto,zhou2023agents,zhou2024agents2,langchain_pandya2023automating} have been introduced to augment LRMs by enabling interaction with external APIs, search engines, and computational tools~\cite{wu2025agenticreasoningreasoningllms}. In December 2024, Google introduced its initial research-oriented implementation, Gemini Deep Research~\cite{google_gemini_deep_research_2025}, focusing on the agentic system with multi-step reasoning and knowledge integration. Building on these advancements, the task of deep research report generation~\cite{DeepResearcher_zheng2025deepresearcher, OpenResearcher_zheng2024openresearcher, multimodal_deepresearcher_yang2025multimodal, liang2025towards} emerged as a specialized benchmark designed to systematically exploit agentic capabilities. Recent works have further extended this to personalized research evaluations~\cite{liang2025towards} and open sandboxes like DeepResearchGym~\cite{coelho2025deepresearchgym}, testing abilities to execute long-horizon planning and synthesize heterogeneous evidence into faithful analyses~\cite{survey2_zhang2025deep, survey1_xu2025comprehensive, survey3_huang2025deep}.

The field is advancing rapidly, driven by proprietary DRAs such as OpenAI Deep Research~\cite{openai_deep_research_2025}, Tongyi Deep Research~\cite{tongyidr, tongyi2025deepresearch}, Grok DeeperSearch~\cite{xai_grok_deepsearch_2025}, and AutoGLM-Research~\cite{zhipu_autoglm_research_2025}. Parallel open-source efforts include OAgents~\cite{zhu-etal-2025-oagents}, Jina-AI node-DeepResearch~\cite{jina_ai_node_deepresearch_2025}, Agentic Reasoning~\cite{wu2025agenticreasoningreasoningllms}, DeepResearcher~\cite{DeepResearcher_zheng2025deepresearcher}, OpenResearcher~\cite{OpenResearcher_zheng2024openresearcher}, and WebLeaper~\cite{tao2025webleaper}. Collectively, these systems center around planning, query formulation, knowledge discovery, and report generation~\cite{survey2_zhang2025deep, survey1_xu2025comprehensive}. 

Mainstream DRAs~\cite{multimodal_deepresearcher_yang2025multimodal, wu2025agenticreasoningreasoningllms} often employ monolithic architectures or specific agentic strategies. For instance, Agentic Reasoning~\cite{wu2025agenticreasoningreasoningllms} incorporates a Mind-Map agent for robust planning, while Flash-Searcher~\cite{qin2025flash} introduces DAG-based parallel execution to enhance information seeking efficiency. In terms of knowledge synthesis, methods range from structure-controlled generation~\cite{AgentLaboratory_schmidgall2025agent} to multimodal reporting~\cite{multimodal_deepresearcher_yang2025multimodal}. Beyond prompt-based agents, recent advances in reinforcement learning~\cite{deepseek_guo2025deepseek, dapo_yu2025dapo, arpo_dong2025agentic} and agent distillation, such as Chain-of-Agents~\cite{li2025chain} and A$^2$FM~\cite{chen2025AFM}, strive to unlock stronger capabilities by optimizing reasoning trajectories and tool-use behaviors end-to-end.

However, a significant gap remains in effectively synthesizing high-quality data to fine-tune open-source models for state-of-the-art deep research. Although TaskCraft~\cite{shi2025taskcraft} has introduced scalable mechanisms for automating complex task generation via width-based extension, and multi-agent distillation works like Chain-of-Agents~\cite{li2025chain} and A$^2$FM~\cite{chen2025AFM} have made strides in transfer learning, they do not fully capture the iterative, exploratory essence of the complete deep research process—from initial query planning to final report synthesis. By explicitly modeling this workflow to generate training corpora, we aim to democratize access to powerful research assistants, reducing dependency on proprietary APIs and enabling autonomous, evidence-based reasoning in open-source models.

\section{Method}

Our objective is to develop an end-to-end, open-source model for complex research tasks that minimizes reliance on intricate prompt engineering and hand-designed workflows. We propose a three-stage training pipeline: (1) Design a power and generation deep research report generation workflow and synthesize high-quality trajectory data generation for Supervised Fine-Tuning (SFT), (2) Conduct SFT on an open source language model to empower it to learn the research process, and (3) Conduct reinforcement learning to further enhance the model's tool-use deep research report generation capacity.  

\subsection{High-Quality Trajectory Generation for SFT}

The foundation of our approach is a meticulously curated dataset of deep research tool-integrated reasoning trajectories.
The process involves two key steps: data synthesis and trajectory synthesis.

\begin{figure}[!th]
    \centering
\includegraphics[width=1\textwidth]
{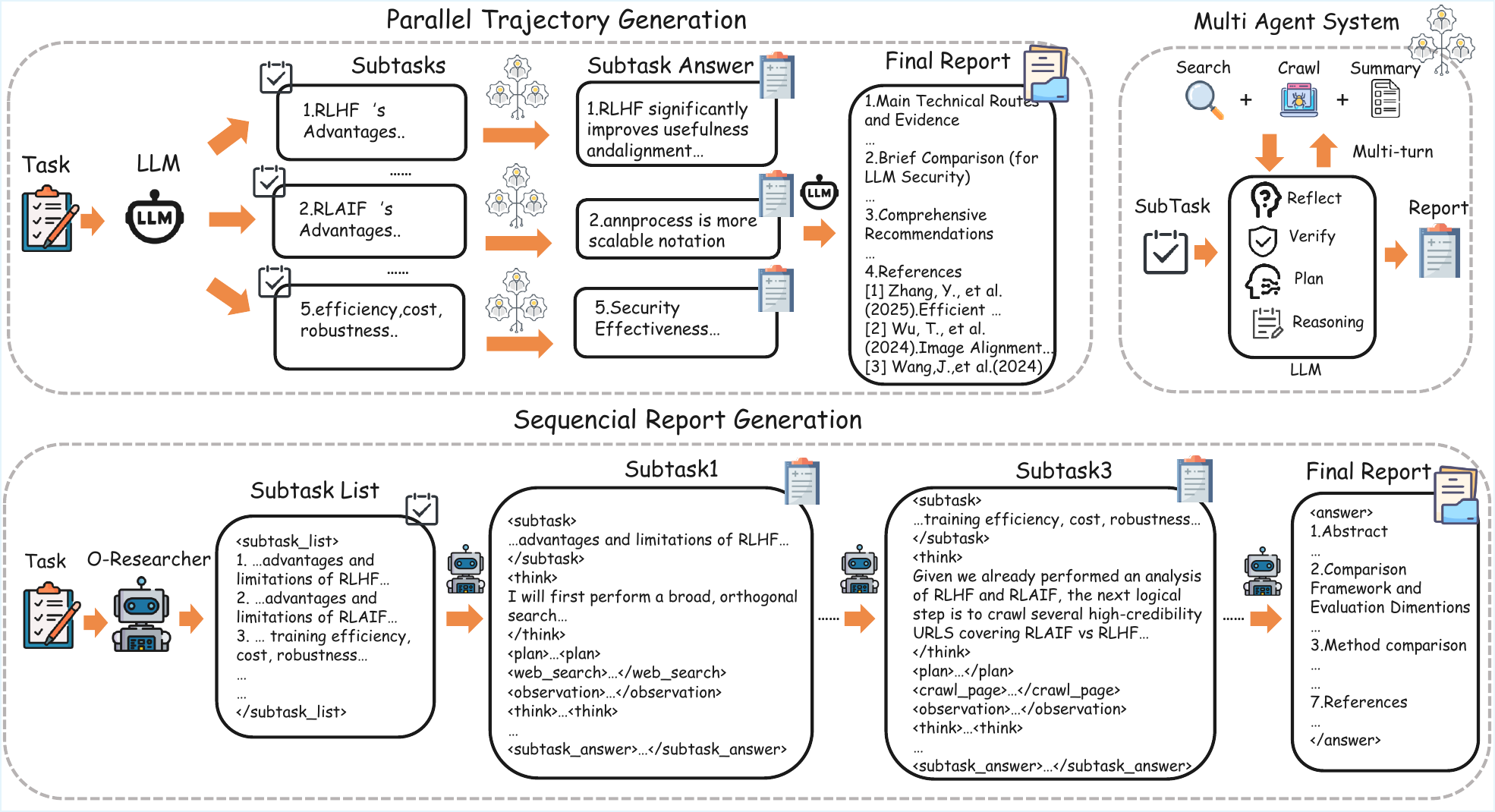}
    \caption{The report generation process of O-Researcher. A query is broken down into multiple sub-queries, which are then independently and parallelized by different agents through tool-integrated reasoning to generate sub-query reports. These sub-query reports are then aggregated through the summarizer agent to generate the final report. All traces and reports of different sub-queries are concatenated as the supervised-training data for this query.}
    \label{method_1}
\end{figure}

\paragraph{Producing Trajectory with Parallel Execution.} 
As depicted in \autoref{method_1}, we designed a parallel execution workflow to generate research reports, simulating the rigorous division of labor in research teams. Drawing on the width-based task extension methodology proposed in TaskCraft~\citep{shi2025taskcraft}, we employ a hierarchical strategy to manage complex queries. A given query is first decomposed by a planner into a set of orthogonal sub-queries, effectively transforming a long-horizon problem into manageable, parallelizable atomic tasks. Different models then independently conduct multiple rounds of tool-integrated reasoning—executing a continuous \textit{Plan-Execute-Observe} loop—to produce detailed sub-query reports. These intermediate reports are then aggregated by a dedicated summarizer model to form the final answer. Meanwhile, the trajectories underlying all sub-reports are collected and merged into a single SFT reasoning trace, ensuring the training data captures the full depth of the parallel reasoning process.


\paragraph{Query Synthesis and Trajectory Collection.} 
To construct a high-quality corpus for supervised fine-tuning, we curated a seed set of 5,000 queries derived from both mature open-source datasets (including Zhihu-KOL~\cite{wangrui6_Zhihu-KOL_2024}, WideSearch~\cite{ByteDance-Seed_WideSearch}, and ELI5~\cite{sentence-transformers_eli5}) and LLM-synthesized topics covering underrepresented domains.
These queries were specifically selected for their open-ended nature and high complexity. 
Following the initial generation by our agentic workflow, we applied rigorous filtering to remove low-quality or short-path trajectories. 
This process ultimately yielded 3,500+ premium instruction-response pairs, which serve as the foundation for our SFT stage.



To construct high-fidelity training data, we employ a diversity-driven synthesis strategy~\cite{li2025chain, chen2025AFM} followed by a rigorous multi-stage rejective sampling pipeline.
For each curated query, our agentic framework—comprising a planner, a tool-user, and a summarizer—first decomposes the query into sub-tasks.  Each sub-task undergoes multi-step\textit{Thought--Tool--Observation} interactions~\cite{qin2025flash}.to generate a corresponding sub-report. Finally, a fusion model consolidates these decomposed sub-reports into a coherent final answer, accompanied by its complete reasoning trace.

\paragraph{Quality Assurance Pipeline}
To ensure the generation of high-fidelity training data, we have designed a comprehensive, multi-stage rejective sampling pipeline. This pipeline acts as a funnel, starting with a diverse set of candidates and progressively filtering them based on increasingly sophisticated criteria, organized into the following stages:

\begin{itemize}
    \item \textbf{Diversity-Driven Generation:} The pipeline begins by generating three distinct candidate trajectories for each query. This initial oversampling is crucial for exploring diverse reasoning paths and increasing the likelihood of producing at least one high-quality output.

    \item \textbf{Rule-Based Hard Rejection:} The generated candidates first undergo a battery of deterministic checks to eliminate structurally flawed or overly simplistic trajectories. A trajectory is immediately rejected if it fails any of the following criteria:
    \begin{itemize}
        \item \textbf{Completeness:} It must contain all necessary tool invocations and correctly closed formatting tags (e.g., \texttt{<web\_search>} and \texttt{</web\_search>}).
        \item \textbf{Context Length:} The total token count must remain within the effective context window ($\le$ 64k tokens).
        \item \textbf{Complexity Thresholds:} To ensure sufficient depth, the trajectory must demonstrate at least 10 reasoning steps and 5 distinct tool-use actions.
        \item \textbf{Consistency:} The output must pass strict format validity checks and maintain language consistency with the input query.
    \end{itemize}
    
    \item \textbf{Model-Based Semantic Filtering:} Trajectories that survive deterministic checks proceed to semantic evaluation by a Qwen3-based LLM-as-a-Judge. This model assesses higher-order qualities beyond rule-based constraints—including logical coherence, tool-use relevance, and evidential grounding—to identify and select the most optimal trajectory for the final dataset.
    
    \item \textbf{Human-in-the-Loop Verification:} As the final layer of validation, the highest-rated trajectories undergo topic-stratified human spot-checking. If a sample is flagged as low-quality, it triggers a regeneration loop, where the original query is re-processed until a valid trajectory is produced and verified.
\end{itemize}

This exhaustive funneling approach ensures that only deep-research trajectories meeting our highest standards are retained for the subsequent SFT and RL phases. Further details are provided in the Appendix.

\paragraph{Structured Data Representation}
To transform raw interaction data into effective training signals, we serialize the trajectories using a coherent XML-style schema. The training sequences are structured to explicitly expose the model's step-wise reasoning and tool-use behaviors, organized into the following tag categories:

\begin{itemize}
    \item \textbf{Workflow Control Tags:} Define the high-level structure of the problem-solving process.
    \begin{itemize}
        \item \texttt{<subtask\_list>}: Decomposes the main query into a sequence of manageable sub-problems.
        \item \texttt{<subtask>}: Marks the beginning of the execution for a specific sub-problem.
    \end{itemize}

    \item \textbf{Cognitive Tags:} Encapsulate the internal reasoning and planning processes.
    \begin{itemize}
        \item \texttt{<think>}: Contains the internal reasoning monologue, analysis, and strategic thinking before an action is taken.
        \item \texttt{<plan>}: Outlines a concrete, step-by-step action plan derived from the preceding thought process.
    \end{itemize}

    \item \textbf{Action Tags:} Represent executable tool calls that the model learns to invoke to gather external information.
    \begin{itemize}
        \item \texttt{<web\_search>}: Invokes the web search tool with a set of queries.
        \item \texttt{<crawl\_page>}: Invokes the web crawler tool with specific URLs.
    \end{itemize}
    
    \item \textbf{Feedback Tag:} Contains the raw output returned by a tool after an action is executed.
    \begin{itemize}
        \item \texttt{<observation>}: Captures the direct results from a tool call (e.g., search snippets, webpage content).
    \end{itemize}

    \item \textbf{Response Tags:} Mark the synthesized conclusions at different stages of the process.
    \begin{itemize}
        \item \texttt{<subtask\_answer>}: Provides the conclusive answer for a single subtask.
        \item \texttt{<suggested\_answer>}: Marks the final, comprehensive report that consolidates all subtask answers.
    \end{itemize}
\end{itemize}

This format forces the model to adhere to a structured \textit{Thinking--Acting--Observing--Answering} loop, which is critical for cultivating robust and verifiable research capabilities. The prompt template is shown as follows.

\begin{tcolorbox}[breakable]

{\normalsize\textbf{Available Functions}}

You may only use the following 8 functions to answer the question.  
Each function must be enclosed within its corresponding tags.

\begin{itemize}[leftmargin=10pt, itemsep=6pt]

\item \textbf{subtask\_list}  
Break the main question into independent subtasks.  
Start with \texttt{<subtask\_list>} and end with \texttt{</subtask\_list>}.

\item \textbf{subtask}  
Marks the specific subtask being executed.  
Start with \texttt{<subtask>} and end with \texttt{</subtask>}.

\item \textbf{think}  
Internal reasoning before plan/tool.  
Start with \texttt{<think>} and end with \texttt{</think>}.

\item \textbf{plan}  
Break the subtask into detailed micro steps.  
Start with \texttt{<plan>} and end with \texttt{</plan>}.

\item \textbf{tool}  
Invoke an external tool.

\item \textbf{observation}  
Holds the tool output.

\item \textbf{subtask\_answer}  
Provide the intermediate answer for a subtask.  
Start with \texttt{<subtask\_answer>} and end with \texttt{</subtask\_answer>}.

\item \textbf{suggested\_answer}  
Integrate all subtask answers into the final solution.
\end{itemize}

{\normalsize\textbf{Available Tools}}

\begin{itemize}[leftmargin=10pt, itemsep=6pt]

\item \textbf{\texttt{<web\_search>}}  
Queries separated by \texttt{|}, append \texttt{\&serp\_num=N}.  
Example:  
\texttt{<web\_search>AI trends | LLM safety\&serp\_num=20</web\_search>}

\item \textbf{\texttt{<crawl\_page>}}  
Fetch deeper information from URLs.

\item You may call \texttt{<web\_search>} multiple times as needed.
\end{itemize}

{\normalsize\textbf{Tool Usage Guide}}

\begin{itemize}[leftmargin=10pt, itemsep=6pt]
\item If retrieved information is irrelevant, refine your queries and repeat \texttt{<web\_search>}.
\item Continue searching until you have high-confidence evidence.
\end{itemize}

{\normalsize\textbf{Trail Notes}}

\begin{itemize}[leftmargin=10pt, itemsep=2pt]

\item \textbf{Workflow:}  
Start with \texttt{<subtask\_list>}.  
Then repeatedly perform:  
\texttt{think → plan → tool → observation},  
until sufficient info is gathered, then generate \texttt{<subtask\_answer>}.

\item \textbf{Information Gathering:}  
Use tools multiple times when necessary.

\item \textbf{Tag Restrictions:}  
Special tags must not appear in free text, especially inside \texttt{<think>}.
\end{itemize}

{\normalsize\textbf{Function Association Instructions}}

\begin{itemize}[leftmargin=10pt, itemsep=2pt]
\item Always begin with \texttt{<subtask\_list>}.
\item Then start the first \texttt{<subtask>}.
\item Inside each \texttt{subtask}, \texttt{<think>} must appear before plan or tool.
\item Output \texttt{<subtask\_answer>} once enough information is gathered.
\item After all subtasks, output \texttt{<suggested\_answer>}.
\end{itemize}

{\normalsize\textbf{Answering Tips}}

\begin{itemize}[leftmargin=10pt, itemsep=2pt]
\item Final \texttt{<suggested\_answer>} must be entirely in English.
\item Must include: Introduction, Body, Conclusion, References.
\item Every key fact must include a citation like \texttt{ [1] }.
\end{itemize}

{\normalsize\textbf{References Section}}

Each reference entry must follow:

\begin{itemize}[leftmargin=10pt, itemsep=2pt]
\item \texttt{[Number]. URL – Webpage Title}
\end{itemize}

\end{tcolorbox}

\subsection{Reinforcement Learning from AI Feedback (RLAIF)}

To further enhance the model's capability to produce high-quality, novel, and comprehensive research reports, we implemented a reinforcement learning stage utilizing Group Relative Policy Optimization (GRPO)~\citep{schulman2017proximal}.

\begin{figure}[!th]
    \centering
\includegraphics[width=1\textwidth]{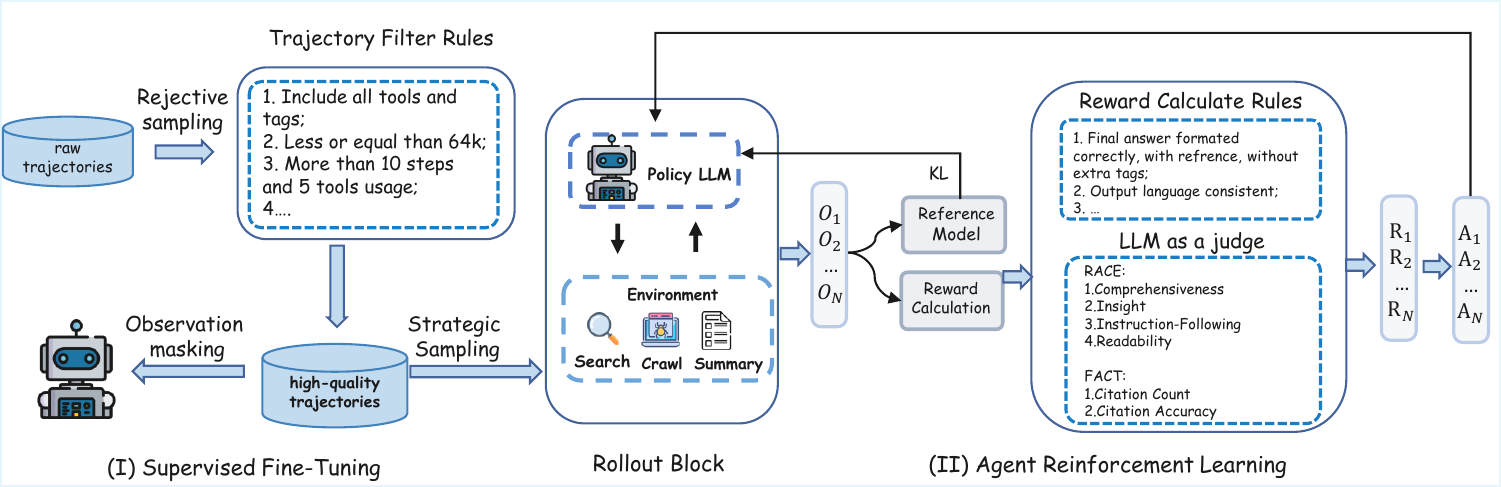}
    \caption{The deep research model training process of our work, which has an SFT stage and an RL stage.}
    \label{method_2}
\end{figure}

\paragraph{Preference Data Curation.}
As depicted in \autoref{method_2}, we began by synthetically generating a diverse set of research questions across multiple domains using an auxiliary large language model (LLM). To construct a preference dataset that is both challenging and informative for reinforcement learning, we filtered these questions based on the performance variance of our supervised fine-tuned (SFT) model. Specifically, for each question, we generated eight distinct responses and evaluated them. Questions that resulted in consistently high scores (indicating trivial difficulty) or consistently low scores (indicating intractable difficulty) were discarded. This filtering approach isolates queries within a "sweet spot" of difficulty, thereby maximizing the learning signal for the policy model during training.

\paragraph{Reward Function Design.}
Our reward function is designed to balance report quality, tool utilization efficiency, and format compliance. It is formulated as a weighted combination of three primary components:
\begin{equation} \label{eq:reward}
R = w_1 R_{\text{base}} + w_2 R_{\text{tool}} + w_3 R_{\text{format}}.
\end{equation}
We set $w_1 = 0.6$, $w_2 = 0.2$ and $w_3 = 0.2$, placing stronger emphasis on high-quality report generation while still encouraging appropriate tool-use behavior.

\paragraph{Base Quality Reward} $(R_{\text{base}})$.
The base quality reward is obtained from an LLM-as-a-Judge that evaluates each (question, generated report) pair along four dimensions: comprehensiveness, insight, instruction-following, and readability. Each dimension is computed via a weighted sum of its criteria, and $R_{\text{base}}$ is the average across dimensions.

\paragraph{Tool-Usage Reward} $(R_{\text{tool}})$.
To encourage appropriate evidence collection, we define
\[
N_{\text{calls}} = \min(\texttt{web\_search}, \texttt{crawl\_page}), 
\qquad
N_{\min}=2,\; N_{\max}=8.
\]
The tool-usage reward is:
\[
R_{\text{tool}} =
\begin{cases}
0, & N_{\text{calls}} < N_{\min},\\[4pt]
-1, & N_{\text{calls}} > N_{\max},\\[4pt]
\displaystyle \frac{N_{\text{calls}} - N_{\min}}{N_{\max} - N_{\min}}, & \text{otherwise}.
\end{cases}
\]

This design rewards reasonable tool usage while penalizing both insufficient and excessive invocation.

\paragraph{Formatting Reward} $(R_{\text{format}})$.
The formatting reward enforces structural correctness. It verifies two strict conditions:  
(1) all XML-style tags must be symmetrically closed;  
(2) the output must contain a \texttt{<suggested\_answer>} tag.  
Violations of either condition yield zero formatting reward.

\paragraph{Final Reward.}
The final composite reward is normalized to $[0,1]$ to provide a stable training signal.

During GRPO training, we observed several notable trends:
\begin{itemize}
    \item a steady increase in average response length, suggesting that the model learned to generate more detailed reports;
    \item a gradual rise in web-search calls and a larger increase in crawl-page calls, indicating deeper evidence gathering;
    \item an initial sharp decline in policy entropy followed by a gradual rebound, reflecting early convergence to effective strategies before exploring more nuanced behaviors.
\end{itemize}
\section{Experiment}
\subsection{Experimental Setup}
\paragraph{Benchmark.}
We evaluate all the artifacts on two benchmarks. One is DeepResearch Bench~\citep{du2025deepresearchbench}. This benchmark comprises 100 doctoral-level research tasks, designed to test advanced reasoning and information synthesis capabilities. The tasks are distributed across four diverse and knowledge-intensive domains: Science \& Technology, Finance \& Business, Software Engineering, and Others. The other is DeepResearchGym~\citep{coelho2025deepresearchgym}, an open-source framework designed to benchmark deep research systems by combining a reproducible search API with a rigorous, multi-dimensional evaluation protocol that assesses alignment with user information needs, factual faithfulness, and report quality.

\paragraph{Baselines.}
To establish a robust comparison, we include several state-of-the-art systems as baselines. These are categorized into two primary groups based on their architecture and accessibility:

\begin{itemize}
    \item \textbf{Deep Research Agents:} This category encompasses advanced agentic systems designed for multi-step, deep information retrieval, typically employing complex planning and tool-use workflows.
    \begin{itemize}
        \item \textit{Proprietary Systems:} We compare against leading closed-source commercial agents, including OpenAI Deep Research \citep{openai_deep_research}, Gemini-2.5-Pro Deep Research \citep{gemini_deep_research}, Perplexity Deep Research \citep{perplexity_deep_research}, and Grok Deeper Search by xAI \citep{grok}.
        \item \textit{Open-source Frameworks:} To benchmark against accessible academic works, we include MiroFlow \citep{2025miroflow} and OAgents \citep{zhu-etal-2025-oagents}. Additionally, we evaluate WebWeaver~\citep{li2025webweaver}.
    \end{itemize}

    \item \textbf{Deep Research Models:} This category includes Large Language Models (LLMs) that are either inherently integrated with search capabilities or specifically fine-tuned for deep research tasks.
    \begin{itemize}
        \item \textit{Proprietary Search-enhanced LLMs:} We evaluate representative commercial search-integrated models. From OpenAI, we consider O3, GPT-4.1, and GPT-5 \citep{openai_models}. We also include the Gemini-2.5 family (Pro and Flash) \citep{gemini25_tech_report}, Perplexity Sonar and Sonar-Pro \citep{perplexity_sonar}, Kimi-K2 \citep{kimi_moonshot}, and MiniMax M2~\citep{minimax2025m2}. These models represent the state-of-the-art in standard retrieval-augmented generation.
        \item \textit{Open-source Models:} To assess the performance of O-Researcher, we compare our SFT and RL variants alongside other state-of-the-art open-weights deep research models, including Tongyi-Deep Research \citep{team2025tongyi} and MiroThinker \citep{team2025mirothinker}.
    \end{itemize}
\end{itemize}

\paragraph{Impact of Context Length.} To investigate the effect of context length on learning complex research workflows, we conducted experiments training on trajectory data truncated to 32k and 64k tokens. Our findings indicate a substantial performance improvement when scaling from 32k to 64k, suggesting that a longer context is crucial for capturing the complete research arc. However, we observed diminishing returns when scaling from 64k to 128k, implying a potential plateau in performance gains from context length alone.

\subsection{Evaluation Metrics}


To provide a multi-faceted assessment of model performance, Deep-Research-Bench~\cite{du2025deepresearchbench} employs two sets of metrics: \textbf{RACE} for evaluating the qualitative aspects of the report and \textbf{FACT} for quantifying its factual correctness and citation quality. These are detailed as follows:

\begin{description}
    \item[\textbf{RACE (Report Quality):}] This metric assesses the overall quality and presentation of the report through four criteria:
    \begin{itemize}
        \item \textbf{Comprehensiveness:} The extent to which the report covers all key aspects of the query.
        \item \textbf{Insight/Depth:} The level of analysis, novelty, and depth beyond simple information aggregation.
        \item \textbf{Instruction-Following:} How well the report adheres to any implicit or explicit constraints in the user's query.
        \item \textbf{Readability:} The clarity, structure, and coherence of the writing.
    \end{itemize}

    \item[\textbf{FACT (Factual Correctness):}] This metric focuses on the reliability and accuracy of the report's content, comprising two main items:
    \begin{itemize}
        \item \textbf{Citation Accuracy:} Whether the provided citations accurately support the claims in the text.
        \item \textbf{Effective Citations:} The relevance and quality of the sources cited to support the report's claims.
    \end{itemize}
\end{description}

\begin{table}[!htb]
  \centering
  \caption{Overall evaluation results of DeepResearch Bench.
  }
  \label{tab:main_comparison}
  \resizebox{0.98\textwidth}{!}{
  \begin{tabular}{l *{7}{c}}
    \toprule
    \multirow{2}{*}{\textbf{Model}}
      & \multicolumn{5}{c}{\textbf{RACE}} 
      & \multicolumn{2}{c}{\textbf{FACT}} \\
    \cmidrule(lr){2-6}\cmidrule(lr){7-8}
      & \textbf{Overall} & \textbf{Comp.} & \textbf{Depth}
      & \textbf{Inst.} & \textbf{Read.}
      & \textbf{C. Acc.} & \textbf{E. Cit.} \\

    \midrule
    \multicolumn{8}{c}{Deep Research Agents} \\
    \midrule
      Grok Deeper Search & 40.24
      & 37.97 & 35.37 & 46.30 & 44.05 & 83.59 & 8.15 \\
      Perplexity Deep Research & 42.25
      & 40.69 & 39.39 & 46.40 & 44.28 & \textbf{90.24} & 31.26 \\
      Gemini-2.5-Pro Deep Research & 48.88
      & 48.53 & 48.50 & 49.18 & \textbf{49.44}
      & 81.44 & 111.21 \\
      O3 Deep Research & 46.98 & 46.87 & 45.25 & 49.27 & 47.14 & 77.96 & 40.79 \\
      MiroFlow & 44.87 & 44.57 & 49.30 & 45.45 & 46.11 & 86.98 & 23.93 \\
      WebWeaver & 50.62 & \textbf{51.29} & 51.00 & 49.98 & 48.89 & 78.25 & \textbf{166.73} \\
       OAgents-DR & \textbf{50.76} & 50.39 & \textbf{51.20} & \textbf{50.32} & 49.41 & 36.98 & 12.56 \\
    \midrule
    \multicolumn{8}{c}{Deep Research Models} \\
    \midrule
      Claude-3-7-Sonnet & 40.67
      & 38.99 & 37.66 & 45.77 & 41.46
      & 93.68 & 32.48 \\
      Claude-3-5-Sonnet  & 28.48
      & 24.82 & 22.82 & 35.12 & 35.08 & \textbf{94.04} & 9.78 \\
      Perplexity-Sonar-Reasoning-Pro & 40.22
      & 37.38 & 36.11 & 45.66 & 44.74
      & 39.36 & 8.35 \\
      Perplexity-Sonar-Pro & 38.93
      & 36.38 & 34.26 & 44.70 & 43.35 & 78.66 & 14.74 \\
      Gemini-2.5-Pro & 35.12
      & 34.06 & 29.79 & 41.67 & 37.16 & 81.81 & \textbf{32.88} \\
      Gemini-2.5-Flash & 32.39
      & 31.63 & 26.73 & 38.82 & 34.48 & 81.92 & 31.08 \\
      OpenAI O3 & 43.71 & 42.02 & 38.80 & 50.29 & 45.90 &  19.00 &  1.10 \\
      GPT-5 & 46.77 & 45.41 & 44.54 & 50.29 & 47.47 & 32.25 & 12.21 \\
      GPT-4.1 & 33.46
      & 29.42 & 25.38 & 42.33 & 40.77 & 54.1 & 5.0 \\
      Kimi-K2-Thinking & 45.65 & 44.57 & 43.94 & 48.97 & 44.65 & 11.39 & 0.97 \\
      Kimi-K2 & 44.47 & 42.78 & 39.65 & \textbf{50.82} & 46.00 &  5.83 &  0.27 \\
      MiniMax M2 & 46.06 & 45.19 & 44.56 & 49.01 & 44.68 & 18.33 & 2.88 \\
    Tongyi-Deep Research & 45.66 & 44.70 & 44.20 & 48.82 & 44.20 & - & - \\
    MiroThinker & 41.79 & 40.22 & 35.41 & 47.79 & 46.64 & 26.32 & 0.22 \\
    Qwen-2.5-72B-Instruct & 33.38 & 30.27 & 23.41 & 43.80 & 39.49 & 44.27 & 8.96 \\     
    O-Researcher-SFT   & $46.24_{\textcolor{gentlegreen}{+12.86\uparrow}}$ & $44.41_{\textcolor{gentlegreen}{+14.14\uparrow}}$ & $46.84_{\textcolor{gentlegreen}{+23.43\uparrow}}$ & $46.79_{\textcolor{gentlegreen}{+2.99\uparrow}}$ & $46.76_{\textcolor{gentlegreen}{+7.27\uparrow}}$ & $29.13_{\textcolor{softred}{-15.14\downarrow}}$ & $22.63_{\textcolor{gentlegreen}{+13.67\uparrow}}$ \\     
    O-Researcher-RL    & $\textbf{48.48}_{\textcolor{gentlegreen}{+15.10\uparrow}}$ & $\textbf{47.32}_{\textcolor{gentlegreen}{+17.05\uparrow}}$ & $\textbf{49.54}_{\textcolor{gentlegreen}{+26.13\uparrow}}$ & $48.64_{\textcolor{gentlegreen}{+4.84\uparrow}}$ & $\textbf{47.58}_{\textcolor{gentlegreen}{+8.09\uparrow}}$ & $31.99_{\textcolor{softred}{-12.28\downarrow}}$ & $26.01_{\textcolor{gentlegreen}{+17.05\uparrow}}$\\
    \bottomrule
  \end{tabular}
  }
  
\end{table}
\subsection{Performance Comparison}
As shown in Table~\ref{tab:main_comparison}, the experimental results demonstrate that O-Researcher-RL establishes a new state-of-the-art among open-weights deep research models, significantly outperforming concurrent open-source works such as Tongyi-Deep Research~\citep{team2025tongyi}\footnote{FACT cannot be calculated as the model fails to provide reference links.} and MiroThinker~\citep{team2025mirothinker}. Furthermore, our model transcends the capabilities of leading search-enhanced commercial LLMs, surpassing both the GPT-5 baseline \citep{openai_models} and OpenAI O3~\citep{openai_models}, while also outperforming specialized proprietary agents like Perplexity Deep Research \citep{perplexity_deep_research}. This indicates that our framework successfully bridges the performance gap between open-source models and top-tier closed-source systems.

The most notable performance transformation is observed when comparing O-Researcher to its backbone, Qwen-2.5-72B-Instruct~\citep{yang2024qwen25}. While the base model maintains a relatively high citation accuracy (44.27\%), it yields a very low volume of effective citations (8.96), reflecting the inherent limitation of general-purpose LLMs in conducting exhaustive evidence-based reports. Through training on deep research trajectories, O-Researcher-SFT learns to execute complex research protocols and generate significantly more comprehensive reports, which is evidenced by a substantial surge in effective citations (+13.67). However, the increased length and complexity of these generated trajectories also introduce a degree of redundancy and noise, leading to a noticeable decline in citation accuracy to 29.13\%.

To address this trade-off, our reinforcement learning stage incorporates the reward function (Eq.~\ref{eq:reward}) specifically designed to optimize report effectiveness and factual grounding. As a result, O-Researcher-RL not only achieves the highest overall RACE score of 48.48 but also successfully mitigates the precision loss observed in the SFT stage. Specifically, the RL variant improves Citation Accuracy from 29.13\% to 31.99\% while further boosting Effective Citations to 26.01. This evolutionary path suggests that our RL strategy effectively filters out redundant information and enforces more rigorous evidence-linking, striking a superior balance between analytical depth and factual rigor. In contrast, other reasoning-centric models such as OpenAI O3 and Kimi-K2-Thinking \citep{kimi_moonshot} exhibit critically low citation metrics, highlighting O-Researcher's unique advantage in providing transparent and verifiable research insights through its optimized training pipeline.

\begin{table}[!th]
\centering
\caption{Evaluations on DeepResearchGym-Commercial-100.}
\label{tab:deepresearchgym}
\resizebox{0.8\textwidth}{!}{
\begin{tabular}{lcccccc}
\toprule
\multirow{2}{*}{Artifacts} & \multicolumn{2}{c}{Relevance} & \multicolumn{2}{c}{Faithfulness} & \multicolumn{2}{c}{Quality} \\ \cline{2-3} \cline{4-5} \cline{6-7} 
& KPR           & KPC           & Precision        & Recall        & Clarity & Insight      \\
\midrule
\multicolumn{7}{c}{Deep Research Agents} \\
\midrule
Perplexity Deep Research & 66.14          & 1.82           & 30.56   & \textbf{98.98}             & 96.3             & 86.4             \\
O3 Deep Research    & 62.4           & 1.56           & 33.46    & 83.14            & 92.8             & 76.8             \\
MiroFlow           & \textbf{80.55} & 1.83           & 25.93   & 84.48             & 96.1             & 95.4             \\
OAgents-DR             & 80.17          & \textbf{1.16}           & \textbf{35.65}      & 97.11          & \textbf{99.5}    & \textbf{99.7}    \\
\midrule
\multicolumn{7}{c}{Deep Research Models} \\
\midrule
Kimi-K2-Thinking    & 64.32          & 1.28           & 10.56   & 87.78             & 97.4             & 82.8             \\
MiniMax M2          & 37.56          & \textbf{0.91}  & 5.81    & 50.57             & 52.6             & 43.1             \\
Tongyi-DeepResearch     & 67.33          & 1.8            & -      & -              & 95.90            & 79.1             \\
O-Researcher-32B    & 68.35          & 0.93           & 32.01      & 94.44          & 95.9             & 91.1             \\
O-Researcher-72B    & \textbf{77.28}          & 1.74           & \textbf{51.45}   & \textbf{94.61}    & \textbf{100.00}           & \textbf{99.3}             \\ \bottomrule
\end{tabular}
}
\end{table}

\subsection{Evaluations on DeepResearchGym}
 In our experimental setup, we adopt the "Commercial" configuration, where each system utilizes its proprietary search infrastructure to retrieve information. We report performance on the official 100-task-subset. Table~\ref{tab:deepresearchgym} summarizes the performance of various artifacts, categorized into search-enhanced LLMs~\citep{kimi_moonshot, minimax2025m2}, deep research agents~\citep{perplexity_sonar, openai_deep_research_2025, 2025miroflow, zhu-etal-2025-oagents}, and open-sourced deep research models~\citep{tongyi2025deepresearch}. Notably, O-Researcher-72B establishes a new state-of-the-art among the evaluated open-source models and remains highly competitive against closed-source commercial agents. In terms of report quality, O-Researcher-72B achieves perfect or near-perfect scores, recording a 100.00 in Clarity and 99.3 in Insight, outperforming commercial baselines such as Sonar-Deep-Research and O3-Deep-Research.
Crucially, regarding retrieval faithfulness, O-Researcher-72B exhibits exceptional grounding. It attains a Citation Precision of 51.45, which is the highest across all categories in the table. This indicates that while maintaining high recall, O-Researcher avoids the trade-off of hallucinated citations often observed in other systems. Furthermore, in terms of relevance (KPR), O-Researcher-72B substantially outperforms standard search-enhanced LLMs and effectively rivals specialized agents, demonstrating a robust ability to synthesize comprehensive reports that satisfy complex user information needs. 
\section{Discussion}

\paragraph{Parallel execution enables effective report generation.} To quantitatively assess the impact of our proposed methodology, we compare the performance of GPT-5 with and without the parallel execution workflow. As illustrated in \autoref{tab:workflow}, employing parallel execution yields a substantial improvement across all evaluated metrics. The overall score increases significantly from 42.92 to 49.60. More specifically, we observe notable gains in Comprehensiveness (40.59 → 49.61) and Insight (38.58 → 48.69), which are critical dimensions for deep research tasks. This demonstrates that decomposing a complex problem into manageable sub-tasks enables the model to conduct a more thorough investigation and generate deeper analysis. Even metrics with already high baseline performance, such as Instruction Following and Readability, see further improvements. These results strongly validate the necessity of a structured workflow for handling complex research queries, as it provides a systematic framework that guides the model to outperform its prompting counterpart consistently.

\begin{table}[htbp!]
\centering
\small
\begin{threeparttable}
\caption{The comparison between GPT-5 with and without parallel workflows.}
\label{tab:workflow}
\sisetup{round-mode=places, round-precision=2, detect-weight=true}
\setlength{\tabcolsep}{10pt}

\begin{tabular}{
    p{3.2cm}  
    *{5}{S[table-format=2.2]}
}
\toprule
Workflows &
{Overall} & {Comp.} & {Ins.} & {Inst.} & {Read.} \\
\midrule
Sequential Execution & 42.92 & 40.59 & 38.58 & 48.05 & 46.88 \\
\textbf{Parallel Execution} & \textbf{49.60} & \textbf{49.61} & \textbf{48.69} & \textbf{50.58} & \textbf{50.32} \\
\bottomrule
\end{tabular}

\end{threeparttable}
\end{table}

\paragraph{Qualitative Analysis of Parallel Execution Workflow.}

\begin{figure}[htbp]
    \centering
    \includegraphics[width=1.0\textwidth]{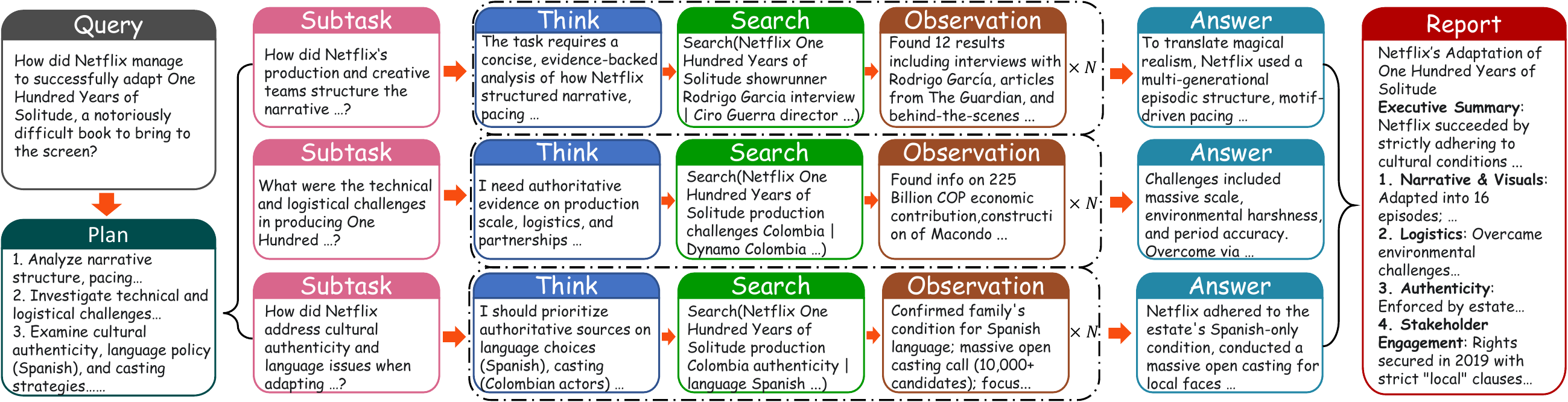} 
    \caption{Qualitative case study of Parallel Execution. The complex adaptation task is decomposed into three distinct subtasks (Narrative, Logistics, Authenticity). Each subtask independently undergoes a \textit{Think-Search-Observation} loop, retrieving high-granularity evidence (e.g., specific budget figures, casting numbers) before being synthesized into a comprehensive report. This contrasts with sequential methods that often fail to maintain such depth across multiple dimensions simultaneously.}
    \label{fig:netflix_workflow}
\end{figure}

To visualize the efficacy of our parallel execution strategy, Figure~\ref{fig:netflix_workflow} presents a case study on the complex adaptation of \textit{One Hundred Years of Solitude}. The workflow demonstrates how O-Researcher decomposes the query into orthogonal dimensions—Narrative Structure, Logistics, and Cultural Authenticity—thereby allowing dedicated agents to execute concurrent \textit{Think-Search-Observation} loops. 

Crucially, this parallelization mitigates the context competition inherent in sequential generation. By isolating the search contexts, the agents successfully retrieve and retain high-granularity evidence, such as the specific "225 Billion COP economic contribution" and "10,000+ casting candidates," which might otherwise be diluted in a single long-context stream. The final aggregation step synthesizes these independent, evidence-rich threads into a cohesive executive summary. This qualitative evidence reinforces our quantitative findings in Table~\ref{tab:workflow}, confirming that structured decomposition is the key driver behind the significant improvements in Comprehensiveness (49.61 vs 40.59) and Insight(48.69 vs 38.58) scores.

\paragraph{Reasoning Steps for Report Generation.}
What is the optimal step number for our report generation workflow? The step number is important since it determines the length of the training data, which directly controls the training cost and the training performance. In this work, we try three different workflow steps: 5, 10, and 20. As illustrated in \autoref{tab:step_ablation}, the results indicate that the 10-step workflow consistently outperforms the 5-step approach across all evaluation metrics, achieving a higher overall score (49.61 vs 48.80). Notably, the 10-step workflow shows the most significant improvement in Comprehensiveness. This suggests that a more fine-grained decomposition of tasks enables a more thorough exploration of the research problem. Therefore, we identify the 10-step workflow as the optimal choice, offering the best balance between performance gains and computational expense. When we increase the step number from 10 to 20, there is no obvious performance gain, but a much longer retrieval context. Therefore, we choose 10 as our step number.

\begin{table}[htbp!]
\centering
\small
\begin{threeparttable}
\caption{Impact of Reasoning Steps. Performance on the OAgents framework}
\label{tab:step_ablation}
\sisetup{round-mode=places, round-precision=2, detect-weight=true}
\setlength{\tabcolsep}{10pt}

\begin{tabular}{
    p{3.2cm}  
    *{5}{S[table-format=2.2]}
}
\toprule
Step Number &
{Overall} & {Comp.} & {Ins.} & {Inst.} & {Read.} \\
\midrule
5 Steps   & 48.80 & 47.89 & 48.31 & 49.81 & 49.91 \\
10 Steps  & 49.61 & 49.60 & 48.71 & \textbf{50.62} & \textbf{50.31} \\
20 Steps  & \textbf{50.76} & \textbf{50.39} & \textbf{51.20} & 50.32 & 49.41 \\
\bottomrule
\end{tabular}

\end{threeparttable}
\end{table}

Our ablation studies reveal two critical factors for enhancing performance in deep research tasks. As shown in Table~\autoref{tab:workflow}, adopting a parallel workflow significantly boosts the overall score of GPT-5 from 42.92 to 49.60. This highlights the importance of structured decomposition for complex queries. Furthermore, Table~\ref{tab:step_ablation} demonstrates a clear correlation between the number of reasoning steps and performance within the OAgents framework. Increasing the steps from 5 to 20 elevates the overall score from 48.80 to a peak of 50.76. This suggests that allowing the model more "thinking time" through a greater number of steps is essential for achieving state-of-the-art results. Both findings underscore that advanced agentic workflows, rather than raw model capability alone, are key to unlocking top-tier performance on these challenging benchmarks.










\section{Conclusion}

In this work, we introduced a novel multi-agent workflow for the automated synthesis of end-to-end deep research data. By structuring collaboration among multiple LLM agents to decompose, debate, and verify complex tasks, this system provides a scalable pipeline for generating high-quality training corpora. Leveraging this synthesized data, we designed a two-stage training strategy that combines supervised fine-tuning with a novel reinforcement learning method to further refine model performance. Extensive experiments on a major deep research benchmark demonstrate that our framework empowers open-source models of various sizes to achieve new state-of-the-art performance, significantly closing the gap with leading closed-source models and establishing new benchmarks in the open-source domain.



\newpage
\section{Contributions}

\textbf{Core Contributors}
\begin{tasks}[
            style=itemize,
            label-width=1em,
            column-sep=3.5em,
            before-skip=1.5ex,
            after-item-skip=1.5ex,
            ](2)
    \task Yi Yao
    \task He Zhu
\end{tasks}

\textbf{Contributors\textsuperscript{*}}
\begin{tasks}[
            style=itemize,
            label-width=1em,
            column-sep=3.5em,
            before-skip=1.5ex,
            after-item-skip=1.5ex 
            ](2)
    \task Piaohong Wang
    \task Jincheng Ren
    \task Xinlong Yang
    \task Qianben Chen
    \task Xiaowan Li
    \task Dingfeng Shi
    \task Jiaxian Li
    \task Qiexiang Wang
    \task Sinuo Wang
    \task Xinpeng Liu
    \task Jiaqi Wu
    \task Minghao Liu
\end{tasks}

\textbf{Corresponding Authors}
\begin{tasks}[
            style=itemize,
            label-width=1em,
            column-sep=3.5em,
            before-skip=1.5ex,
            after-item-skip=1.5ex 
            ](2)
    \task Wangchunshu Zhou
\end{tasks}


\clearpage
\bibliography{index}
\bibliographystyle{unsrt}

\newpage
\appendix

\section{Case Study}

\subsection{Question}

\begin{tcolorbox}[breakable]
What are the current mainstream AI alignment technologies?  
Please briefly compare their performance in terms of safety when handling large-scale language models.
\end{tcolorbox}

\subsubsection{Result From GPT-5:}

\begin{tcolorbox}[breakable]

\subsection*{Main Technical Routes and Evidence}

\begin{itemize}
    \item \textbf{Supervised Fine-tuning (SFT, including instruction fine-tuning and self-instruction)}
        \begin{itemize}
            \item \textbf{Idea:} 
            \item \textbf{Security Performance:} It can significantly improve "usability/compliance with instructions," but in terms of security, it usually only covers risk types clearly present in the training corpus. Its generalization and adversarial robustness are limited, and it is easily broken by jailbreak prompts. While common as a starting point, it often requires subsequent preference optimization or security filtering. Self-instruction work shows that instruction fine-tuning can improve task compliance and generality, but does not solve the problem of adversarial jailbreaks \cite{self_instruct}. Systematic safety assessments (like TruthfulQA and HELM) also suggest that SFT alone has significant room for improvement in "factuality/realism" and "avoidance of inappropriate content" \cite{truthfulqa, helm}.
        \end{itemize}

    \item \textbf{Reinforcement Learning from Human Feedback (RLHF)}
        \begin{itemize}
            \item \textbf{Idea:} Collect human preference comparisons (labeling the outputs of two models as superior or inferior), train a reward model, and then use an RL algorithm like PPO to optimize the policy so that the model's output better aligns with human preferences and safety criteria.
            \item \textbf{Security Performance:} InstructGPT showed that RLHF significantly improves usefulness and alignment, with a measurable reduction in toxicity and inappropriate content \cite{instructgpt}. Anthropic's "Helpful and Harmless Assistant" also proved that RLHF can improve both targets simultaneously, with a lower incidence of harmful responses compared to an SFT-only baseline \cite{anthropic_rlhf}. Limitations include high cost (human annotation) and potential for over-rejection or being bypassed by jailbreak attacks if reward modeling is inappropriate \cite{jailbreak_llms}.
        \end{itemize}

    \item \textbf{Reinforcement Learning from AI Feedback (RLAIF) and "Constitutional AI (CAI)"}
        \begin{itemize}
            \item \textbf{Idea:} Use a set of public principles ("constitution") to constrain the model, then use the model itself or an auxiliary model to evaluate preferences, reducing reliance on human annotation. It is essentially AI feedback replacing part of human feedback.
            \item \textbf{Security Performance:} Anthropic's CAI shows that principle-driven AI feedback can achieve similar or even better performance than human feedback in "harmlessness," and the annotation process is more scalable. The model's rejections are more consistent and traceable to the principles \cite{cai}. This approach is better than pure RLHF in cost and scalability, but also faces challenges from jailbreaks \cite{jailbreak_llms}.
        \end{itemize}

    \item \textbf{Direct Preference Optimization (DPO)}
        \begin{itemize}
            \item \textbf{Idea:} Instead of RL, it uses pairwise preference samples for direct parameter optimization (increasing the log-likelihood of "better" outputs and decreasing that of "worse" ones). It achieves an alignment effect similar to RLHF but with simpler, more stable, and offline training.
            \item \textbf{Security Performance:} The DPO paper shows it is comparable to or even better than RLHF in multiple preference alignment tasks with lower complexity. For security, its capability depends on whether the preference data covers the "harmlessness" dimension. If so, DPO can inherit these security attributes effectively \cite{dpo}. However, it is still subject to data distribution limitations and adversarial attacks.
        \end{itemize}

    \item \textbf{Security Classifiers/Guardrails}
        \begin{itemize}
            \item \textbf{Idea:} Add a security classifier and policy filtering (e.g., intercepting self-harm, illegal guidance, hate speech) to the model's input or output. This can also be implemented via security system prompts or policy templates.
            \item \textbf{Security Performance:} Meta's Llama Guard encodes security policies into a multi-label classifier for auditing inputs/outputs, significantly improving coverage of inappropriate content and easing integration with enterprise policies \cite{llama_guard}. These guardrails are effective for known risk categories but face risks of false positives/negatives and need continuous updates.
        \end{itemize}

    \item \textbf{Red Teaming and Adversarial Training}
        \begin{itemize}
            \item \textbf{Idea:} Systematically design jailbreak/adversarial prompts to stress-test the model, using the findings for data augmentation and iterative training to repair vulnerabilities.
            \item \textbf{Security Performance:} Recent work shows that even LLMs aligned by RLHF/CAI can be jailbroken with a high success rate by automated adversarial methods, demonstrating that existing alignments are not indestructible. Continuous investment in red teaming and defense mechanisms is necessary \cite{jailbreak_llms}.
        \end{itemize}
\end{itemize}

\subsection*{Brief Comparison (for LLM Security)}
\begin{itemize}
    \item \textbf{Security Effectiveness (Preventing Inappropriate/Dangerous Outputs)}
    \begin{itemize}
        \item RLHF and CAI are generally superior to SFT alone, significantly reducing toxicity and providing more consistent rejections \cite{instructgpt, anthropic_rlhf, cai}.
        \item DPO, if using data with security preferences, can achieve a similar level of "harmlessness" as RLHF with simpler training \cite{dpo}.
        \item Security guardrails (like Llama Guard) have strong interception capabilities for known risk types and are a necessary layer for deployment, but do not replace the alignment of the model itself \cite{llama_guard}.
    \end{itemize}

    \item \textbf{Cost and Scalability}
    \begin{itemize}
        \item \textbf{RLHF:} High cost of human annotation and slow iteration \cite{instructgpt, anthropic_rlhf}.
        \item \textbf{CAI/RLAIF:} Relying on AI feedback, the annotation cost is significantly reduced and scalability is better \cite{cai}.
        \item \textbf{DPO:} Simpler, offline training pipeline leads to lower cost; the key is obtaining high-quality preference data \cite{dpo}.
        \item \textbf{Guardrails:} Deployment and maintenance costs are manageable, but policies need continuous updates \cite{llama_guard}.
    \end{itemize}

    \item \textbf{Robustness and Limitations}
    \begin{itemize}
        \item Existing alignment methods (SFT/RLHF/CAI/DPO) can be bypassed by carefully designed prompts; research has demonstrated cross-model transferable jailbreak attacks \cite{jailbreak_llms}. They must be used with red teaming and multi-layered protection.
    \end{itemize}
        
    \item \textbf{Balance between Over-rejection and Practicality}
    \begin{itemize}
        \item If RLHF/CAI overemphasizes "harmlessness," it may result in "over-conservatism." A balance must be achieved through careful reward design and principal trade-offs \cite{anthropic_rlhf, cai}.
    \end{itemize}

    \item \textbf{Authenticity and Factuality (as a security-related dimension)}
    \begin{itemize}
        \item RLHF (like InstructGPT) outperforms unaligned baselines on truthfulness benchmarks like TruthfulQA. However, "hallucination" is not completely eradicated, requiring further enhancements like search augmentation and fact-checking \cite{instructgpt, truthfulqa, helm}.
    \end{itemize}
\end{itemize}

\subsection*{Comprehensive Recommendations}
\begin{itemize}
    \item A common combination in production is: \textbf{SFT as a foundation} + \textbf{preference optimization (RLHF/CAI or DPO)} to improve "helpfulness and harmlessness" + \textbf{security guardrails} for input/output interception + \textbf{continuous red teaming} and adversarial data augmentation.
    \item If cost and scalability are primary considerations, \textbf{CAI/RLAIF} or \textbf{DPO} should be prioritized. If the highest quality is the goal, \textbf{RLHF} remains a strong choice, provided human annotation resources are available. Regardless of the method, continuous security assessments (e.g., TruthfulQA, HELM) and red teaming are essential to maintain security against real-world threats.
\end{itemize}


\end{tcolorbox}



\subsubsection{Result From O-Researcher:}

\begin{tcolorbox}[breakable]

\subsection*{Abstract}
\begin{quote} 
This paper synthesizes reports from multiple subtasks, systematically compares and summarizes the advantages, limitations, and key practical points of mainstream alignment techniques in improving the security of large-scale language models (LLMs), and provides the core principles and representative methods of each technique. Key conclusions include: No single method can comprehensively excel in the dimensions of "usefulness, security, robustness, and cost"; engineering practices often adopt a multi-layer alignment stack (data governance $\rightarrow$ SFT/instruction fine-tuning $\rightarrow$ preference alignment $\rightarrow$ red team/runtime protection) to achieve a compromise \cite{instructgpt, dpo, cai, anthropic_redteam}. RLHF is effective but costly and lacks robustness \cite{instructgpt, cai, anthropic_redteam}. RLAIF significantly reduces costs but is sensitive to the quality of the AI "teacher" and remains fragile against strong adversarial attacks \cite{rlaif_vs_rlhf, critical_eval_ai_feedback, cost_constrained_rlhf}. Policy regularization methods (DPO, etc.) have advantages in sample efficiency but limited jailbreak resilience \cite{efficient_safety_alignment, cdpo, pmlr, clip_misalignment}. This paper provides a structured comparison across multiple dimensions and concludes with practical recommendations for building secure and aligned LLMs.
\end{quote}

\subsection*{Report Structure}
\begin{enumerate}
    \item Comparative Framework and Evaluation Dimensions
    \item Method Comparison (Item-by-Item)
    \item List of Mainstream Alignment Techniques
    \item Practical Recommendations and Multi-Layer Alignment Stack
    \item Conclusions and Research Gaps
    \item References
\end{enumerate}

\subsection*{1. Comparative Framework and Evaluation Dimensions}
For ease of comparison, the following unified evaluation dimensions are adopted:
    \begin{itemize}
        \item \textbf{Direct Safety Effects} (Toxicity, Harmlessness, Bias, Factuality)
        \item \textbf{Usefulness/Helpfulness} (instruction-following, helpfulness metrics)
        \item \textbf{Training efficiency} (data requirements, annotation cost, computation/wall time)
        \item \textbf{Deployment cost} (latency, throughput, model size, inference overhead)
        \item \textbf{Robustness} (red team, jailbreak, hint injection, adversarial examples)
        \item \textbf{Cross-domain/multi-turn dialogue security} (round-cumulative risk, toxicity drift)
        \item \textbf{Governability and auditability} (interpretability, reproducibility)
    \end{itemize}
\textit{(Note: Different studies use different benchmarks. This article cites specific reports to indicate comparability conditions \cite{cai, cai_reused, cost_constrained_rlhf, pmlr}.)}

\subsection*{2. Method Comparison (Item by Item)}


\subsubsection*{\hspace{1.5em} 2.1 Comparison of RLHF with Other Alignment Techniques}
\begin{itemize}
    \item \textbf{Advantages (Security-Oriented)}
        \begin{itemize}
            \item Directly optimizes to human preferences, typically reducing toxicity and improving instruction compliance \cite{instructgpt, cai}.
            \item Mature Engineering: Data pipelines, PPO training, and RM training processes have been systematized in the industry \cite{instructgpt, cai}.
        \end{itemize}
    \item \textbf{Limitations (Security-Oriented)}
        \begin{itemize}
            \item Label consistency and cultural bias issues; RM training may amplify biases or be affected by reward gaming \cite{instructgpt, cai}.
            \item Alignment Tax: Security improvements often come at the expense of downstream task performance \cite{instructgpt}.
            \item Red team evaluations show that training-period alignment alone cannot completely prevent systematic abuse \cite{anthropic_redteam, pmlr}.
            \item High Cost: High-quality human preference labeling is expensive and difficult to scale \cite{cai}.
        \end{itemize}
    \item \textbf{Key Comparisons}
        \begin{itemize}
            \item \textbf{vs. DPO:} Simpler and more efficient training with results close to RLHF, but with a risk of bias in out-of-distribution safety \cite{dpo, cai}.
            \item \textbf{vs. Constitutional AI (CAI):} Achieves large-scale harmless training via AI self-annotation, but its effectiveness is limited by the accuracy of the constitution \cite{dpo, anthropic_redteam}.
            \item \textbf{vs. SFT:} Lower cost and easier to deploy, but with limited safety improvement. Often used as a prerequisite for RLHF/DPO/CAI \cite{instructgpt, cai}.
        \end{itemize}
\end{itemize}

\subsubsection*{\hspace{1.5em} 2.2 Comparison of RLAIF with Other Alignment Techniques}
\begin{itemize}
    \item \textbf{Advantages (Security-Oriented)}
        \begin{itemize}
            \item Using an AI to generate preference labels or scores significantly reduces cost and makes alignment scalable, approaching the effect of RLHF \cite{rlaif_vs_rlhf, instructgpt}.
            \item d-RLAIF (Online LLM Scoring) eliminates the RM training step, simplifying the process \cite{instructgpt}.
            \item Frameworks like SRPO can improve robustness under unstable preferences \cite{srpo}.
        \end{itemize}
    \item \textbf{Limitations (Security-Oriented)}
        \begin{itemize}
            \item Highly sensitive to the quality of the "evaluator/teacher" model \cite{dpo, rlaif_vs_rlhf}.
            \item Robustness to red team/jailbreak evaluation is not significantly better than other training-period alignment methods \cite{pmlr, cost_constrained_rlhf}.
            \item AI-generated labels have poor interpretability and auditability \cite{anthropic_redteam}.
        \end{itemize}
    \item \textbf{Key Comparisons}
        \begin{itemize}
            \item \textbf{vs. RLHF:} Alignment quality is comparable but at a lower cost; however, it requires a strong teacher model \cite{rlaif_vs_rlhf, dpo}.
            \item \textbf{vs. DPO:} d-RLAIF avoids training an RM but online invocation incurs deployment costs; C-DPO provides a compromise \cite{instructgpt, cai}.
            \item \textbf{vs. SFT:} RLAIF can impose more explicit preference targets but relies on AI commenting quality \cite{dpo}.
        \end{itemize}
\end{itemize}

\subsubsection*{\hspace{1.5em} 2.3 Four-dimensional comparison of RLAIF / RLHF / Policy Regularization}
\begin{itemize}
    \item \textbf{Training efficiency:} RLAIF $>$ DPO $\approx$ RLHF \cite{instructgpt, dpo, rlaif_vs_rlhf}.
    \item \textbf{Deployment cost:} DPO $\approx$ SFT is lowest; RLAIF increases cost if online scoring is used \cite{instructgpt, dpo, gemma_j}.
    \item \textbf{Robustness (Toxicity/Jailbreak):} All methods are still vulnerable to systematic jailbreaks. DPO needs specific improvements for OOD security \cite{instructgpt, dpo, anthropic_redteam, srpo, cost_constrained_rlhf}.
    \item \textbf{Multi-turn Conversation Security:} Multi-turn toxicity drift is an open problem for all methods, requiring specific multi-turn alignment data \cite{cai, multiturn_safety}.
\end{itemize}

\subsubsection*{\hspace{1.5em} 2.4 Comparison of Policy Regularization (CLIP/CoT) with Other Alignment Methods}
\begin{itemize}
    \item \textbf{Advantages}
        \begin{itemize}
            \item High sample and data efficiency; can significantly reduce online sampling computation \cite{efficient_safety_alignment, instructgpt}.
            \item CoT alignment can improve inference robustness and cross-task capability \cite{real_toxicity_prompts_arxiv, reasoners_alignment}.
        \end{itemize}
    \item \textbf{Limitations}
        \begin{itemize}
            \item External rewards (like CLIP) are not entirely consistent with text security goals, causing side effects like bias amplification \cite{clip_misalignment, reasoners_alignment}.
            \item Limited performance in automated red team/jailbreak evaluations \cite{anthropic_redteam}.
        \end{itemize}
\end{itemize}

\subsection*{3. List of Mainstream Alignment Techniques}
\begin{itemize}
    \item \textbf{A. Instruction Fine-tuning and Preference Alignment:} Core methods to align LLMs to user intent.
    \begin{itemize}
        \item \textit{Methods:} InstructGPT (SFT$\rightarrow$RM$\rightarrow$PPO) \cite{cai}, DPO \cite{anthropic_redteam}, Constitutional AI \cite{cai_reused}.
    \end{itemize}
    \item \textbf{B. RLHF and its Safe Variations:} Introduce safety constraints into the RLHF process.
    \begin{itemize}
        \item \textit{Methods:} Safe RLHF \cite{safe_rlhf}, Cost-constrained RLHF \cite{cost_constrained_rlhf}, PMLR \cite{pmlr}.
    \end{itemize}
    \item \textbf{C. Data and Post-Training Layers:} Reduce risks via data construction or post-training filtering.
    \begin{itemize}
        \item \textit{Methods:} Toxicity fine-tuning, bias classifiers, debiased datasets \cite{instructgpt, rlaif_vs_rlhf}.
    \end{itemize}
    \item \textbf{D. Security During Inference (Guardrails):} Apply rules and filtering to the decoding chain.
    \begin{itemize}
        \item \textit{Methods:} Guardrail frameworks, restricted decoding, online toxicity classifiers \cite{instructgpt}.
    \end{itemize}
    \item \textbf{E. Automated/Proxy Alignment Paradigm:} Replace human feedback with model self-criticism.
    \begin{itemize}
        \item \textit{Methods:} H1–H5, RLAIF, IDA \cite{dpo}.
    \end{itemize}
    \item \textbf{F. Multimodal and Robust Security:} Attack/defense systems for adversarial prompts.
    \begin{itemize}
        \item \textit{Reviews:} PromptBench and JailbreakBench \cite{rlaif_vs_rlhf}.
    \end{itemize}
\end{itemize}

\subsection*{4. Practical Recommendations and Multi-Layer Alignment Stack}
\begin{itemize}
    \item \textbf{Multi-layer alignment stack:}
        \begin{enumerate}
            \item \textbf{Data Governance:} Data depollution, debiasing, and adversarial pollution detection \cite{instructgpt}.
            \item \textbf{Training Phase:} SFT $\rightarrow$ Preference Alignment (RLHF / DPO / RLAIF / CAI), adding safety constraints as needed \cite{cai, anthropic_redteam, rlaif_vs_rlhf, safe_rlhf, pmlr}.
            \item \textbf{Model Evaluation:} Multi-benchmark evaluation (helpfulness, toxicity, jailbreak) plus red team testing \cite{cost_constrained_rlhf, gemma_j}.
            \item \textbf{Runtime Protection:} Guardrails, decoding filtering, content review, and continuous online monitoring \cite{instructgpt, anthropic_redteam, cost_constrained_rlhf}.
        \end{enumerate}
    \item \textbf{Method Combinations and Trade-offs:} Prioritize DPO/CAI/RLAIF for computational constraints; prioritize RLHF for human preference consistency. Employ multi-layered protection in safety-critical scenarios \cite{instructgpt, dpo, cai, anthropic_redteam, cost_constrained_rlhf}.
    \item \textbf{Risk Management and Governance:} Establish model cards, red team reports, versioning, and external compliance reviews \cite{instructgpt, dpo}.
\end{itemize}

\subsection*{5. Conclusions and Research Gaps}
\begin{itemize}
    \item \textbf{Key Conclusions}
        \begin{itemize}
            \item No single method excels across "usefulness—safety—robustness—cost." Engineering often employs multi-layered, hybrid approaches \cite{instructgpt, dpo, cai, anthropic_redteam}.
            \item RLAIF and DPO offer scalable pathways, but all methods still have shortcomings against strong adversarial attacks and multi-round toxic drift \cite{rlaif_vs_rlhf, dpo, cai, anthropic_redteam, dpo_vs_ppo_study}.
            \item Automated alignment (H1–H5) is an important future direction but faces challenges like reward hacking and error accumulation \cite{dpo}.
        \end{itemize}
    \item \textbf{Research Gaps and Suggested Directions}
        \begin{itemize}
            \item Data and benchmarks for multi-round toxic drift are still insufficient; continuous investment in red teams and benchmarks is needed \cite{cost_constrained_rlhf, gemma_j}.
            \item Provable trade-offs between security, capability, and computation remain unclear. Robust alignment methods across languages and cultures are lacking \cite{instructgpt, dpo}.
            \item Reward model robustness, reward hacking detection, and theoretical guarantees for RLHF/RLAIF are important research directions \cite{dpo, safe_rlhf, pmlr}.
        \end{itemize}
\end{itemize}

\end{tcolorbox}
\end{document}